\documentclass[runningheads]{llncs}
\usepackage[T1]{fontenc}
\usepackage{graphicx,verbatim}
\begin{document}
\title{A Lightweight Large Vision-language Model for Multimodal Medical Images}

\author{Belal Alsinglawi$^1$, Chris McCarthy$^1$, Sara Webb$^1$, Christopher Fluke$^1$, Navid Toosy Saidy$^2$}
\authorrunning{Alsinglawi et al.}
\institute{$^1$Swinburne University of Technology, Melbourne, Australia\\
$^2$PropelHealthAI, Brisbane, Australia\\
\email{cdmccarthy@swin.edu.au}}

\maketitle

\begin{abstract}
Medical Visual Question Answering (VQA) enhances clinical decision-making by enabling systems to interpret medical images and answer clinical queries. However, developing efficient, high-performance VQA models is challenging due to the complexity of medical imagery and diverse modalities. In this paper, we introduce a lightweight, multimodal VQA model integrating BiomedCLIP for image feature extraction and LLaMA-3 for text processing. Designed for medical VQA tasks, our model achieves state-of-the-art performance on the OmniMedVQA dataset. With approximately 8 billion parameters, it requires only two NVIDIA 40 GB A100 GPUs, demonstrating superior efficiency over larger models. Our results show 73.4\% accuracy for open-end questions, surpassing existing models and validating its potential for real-world medical applications. Key contributions include a specialized multimodal VQA model, a resource-efficient architecture, and strong performance in answering open-ended clinical questions.

\keywords{Visual Question Answering  \and Lightweight \and Radiology.}

\end{abstract}
\section{Introduction}
Medical Visual Question Answering (VQA) aims to develop systems that interpret medical images and provide accurate answers to clinically relevant questions \cite{litjens2017survey,esteva2019guide}. Such systems can assist healthcare professionals in diagnosis, treatment planning, and patient education by extracting reliable information from medical images~\cite{shen2017deep}. However, developing effective medical VQA models is challenging due to the complexity of medical images, the need for domain-specific knowledge, and the diversity of imaging modalities~\cite{kermany2018identifying,holzinger2019causability}.

Existing medical VQA models often focus on modality- or illness-specific datasets, such as VQA-RAD~\cite{lau2018dataset} and SLAKE~\cite{liu2021slake}, which are limited in size and scope. These models rely on heavy computational resources and struggle to generalize across medical image modalities~\cite{zhou2020unified}. Recent efforts have aimed to leverage large-scale pretrained models to enhance performance. BiomedCLIP~\cite{zhang2023biomedclip}, a multimodal biomedical foundation model pretrained on fifteen million scientific image-text pairs, has shown significant promise in capturing the nuances of medical imagery. Similarly, large language models like LLaMA~\cite{touvron2023llama} have demonstrated advanced language understanding capabilities that can be beneficial for generating accurate and contextually appropriate answers in medical VQA tasks.However, integrating these models often results in computationally intensive systems, limiting their clinical applicability~\cite{chen2024huatuogpt}.

In this paper, we introduce a lightweight multimodal VQA model for medical imaging, integrating BiomedCLIP for image feature extraction and LLaMA-3 for natural language processing. Our model optimizes the architecture to reduce parameters and computational overhead while maintaining accuracy. Evaluated on the OmniMedVQA dataset \cite{hu2024omnimedvqa}, our model achieves state-of-the-art performance, surpassing existing models in accuracy and efficiency.

Key contributions include:
\begin{itemize}
    \item \textbf{Specialized multimodal VQA model:}  BiomedCLIP and LLaMA-3 are combined for precise image and text processing in medical contexts.
    \item \textbf{Lightweight architecture:}  Approximately 8 billion parameters are used, reducing computational demands while maintaining high performance.
    \item \textbf{State-of-the-art performance:}  Advanced accuracy is achieved on the OmniMedVQA dataset across diverse medical imaging modalities.
    \item \textbf{Open-ended question support:} Handles dynamic clinical tasks by answering open-ended questions, unlike models limited to closed-form questions.
\end{itemize}

Our experiments demonstrate that this model provides a practical and efficient basis for real-world medical VQA applications.

\section{Related Work}
Medical VQA has gained attention for its potential to assist clinicians in interpreting medical images and making diagnostic decisions. Early works focused on specialized datasets and models to address challenges like scarce annotated data and domain-specific knowledge~\cite{XIE2021101985}. The VQA-Med dataset~\cite{ben2019vqa} established a benchmark for medical image understanding and question answering. Hybrid Deep Neural Networks~\cite{harzig2018visual} combined convolutional and recurrent layers to capture spatial and sequential information from images and questions.

Contrastive learning methods like ConVIRT \cite{zhang2022contrastive} and GLoRIA \cite{huang2021gloria} improved medical image representation by leveraging paired images and text. Large-scale pretrained models, such as VisualBERT~\cite{li2019visualbert} and UNITER~\cite{chen2020uniter}, integrated textual and visual information using transformers but struggled with medical domain-specific concepts.

BiomedCLIP~\cite{zhang2023biomedclip}, a multimodal biomedical foundation model, adapted the CLIP architecture~\cite{radford2021learning} for biomedical tasks using PubMedBERT~\cite{gu2021domain} for text and Vision Transformers~\cite{dosovitskiy2020image} for images. It outperformed general-domain models and PubMedCLIP~\cite{singh2024medical} in tasks like image-text retrieval and classification.

Large language models (LLMs) like LLaMA~\cite{touvron2023llama} have advanced text understanding and generation. HuatuoGPT-Vision~\cite{chen2024huatuogpt} integrated medical visual knowledge into multimodal LLMs, using GPT-4V~\cite{achiam2023gpt} to refine medical image-text pairs. Despite progress, models like LLaVA-Med~\cite{li2024llava} face challenges in aligning visual and textual modalities due to data limitations.

Our work builds on these advancements by integrating BiomedCLIP and LLaMA-3 for medical VQA. We fine-tune both components on specialized datasets to align visual and textual representations effectively. By handling higher-resolution images and longer textual descriptions, we aim to improve generalization across medical imaging modalities and tasks.

\section{Methodology}
\subsection{Model Architecture}
\begin{figure}
\centerline{\includegraphics[width=\textwidth]{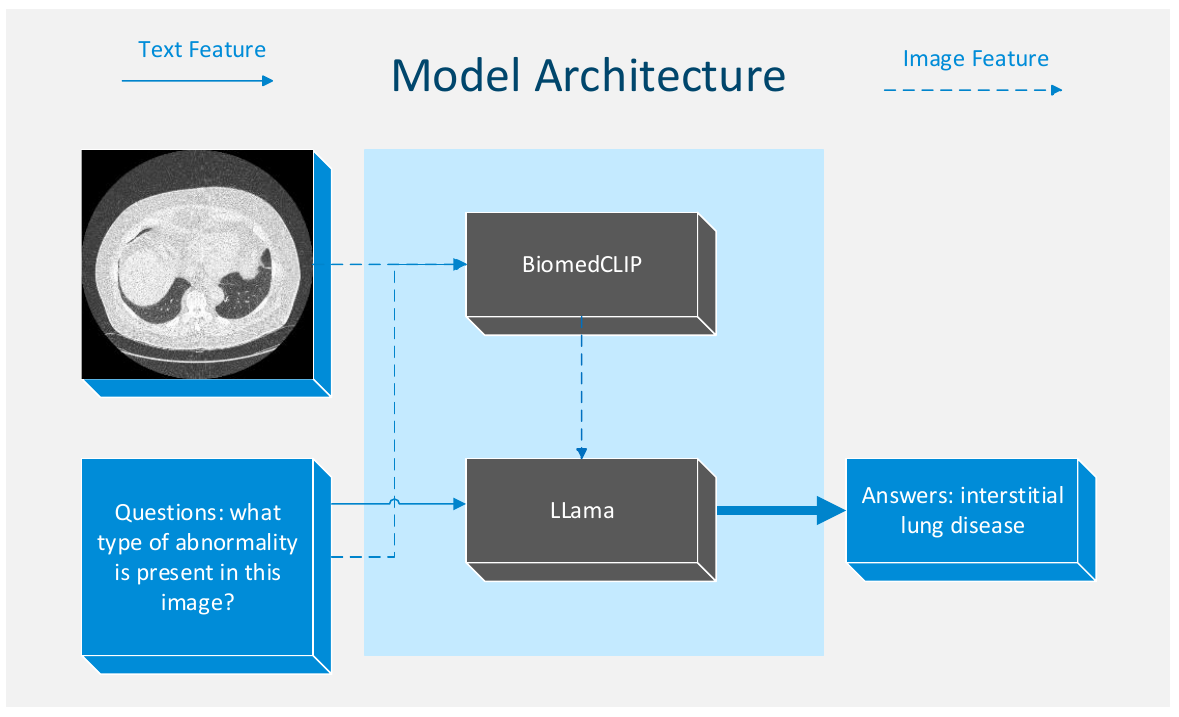}}
\caption{The Architecture of LLama-CLIP model. The model takes an image (left) and an open-ended question, such as "What type of abnormality is present in this image?" The BiomedCLIP module processes the image to generate image features, while LLama encodes the question to extract text features. LLama integrates features and generates the final answer—here, identifying "interstitial lung disease" as the abnormality shown in the image.}
\label{Architecture}
\end{figure}
Our model leverages a hybrid architecture, combining BiomedCLIP and LLama3, and is specifically designed for medical tasks.

As illustrated in Figure~\ref{Architecture}, BiomedCLIP is employed as the image encoder, extracting rich image features from medical images like computed tomography (CT) scans, X-ray images and magnetic resonance imaging (MRI) data. BiomedCLIP reads both the image and the accompanying question to derive meaningful visual representations. Meanwhile, LLama3 acts as the text encoder, converting input questions from the VQA task into high-dimensional text embeddings.

After encoding both image and text features, LLama3 also serves as the feature fusion mechanism, integrating these features to form a combined representation. Finally, the generation module takes this fused representation to generate answers to the given medical questions. This architecture ensures both visual and textual information is effectively captured and utilized to provide responses.

\subsection{Training Process}
The training process of the model is divided into two stages:
\begin{itemize}
    \item[1.] BiomedCLIP is trained independently on a subset of the open-source portion of the OmniMedVQA dataset, focusing on extracting high-quality visual features from medical images. Meanwhile, LLama3-8B is fine-tuned using LoRA (Low-Rank Adaptation), which allows efficient training with reduced computational costs. 
    \item[2.] Once both the image and text encoders are trained separately, the two components are aligned, and a joint fine-tuning process is performed.
\end{itemize}

This final phase ensures that the visual and textual features are well integrated, enhancing the model’s ability to answer medical VQA tasks accurately and efficiently. The combination of these steps ensures optimal model performance while maintaining computational efficiency.

\section{Experimental Setup}
\subsection{Datasets}

\begin{figure}
    \centering
    \includegraphics[width=\linewidth]{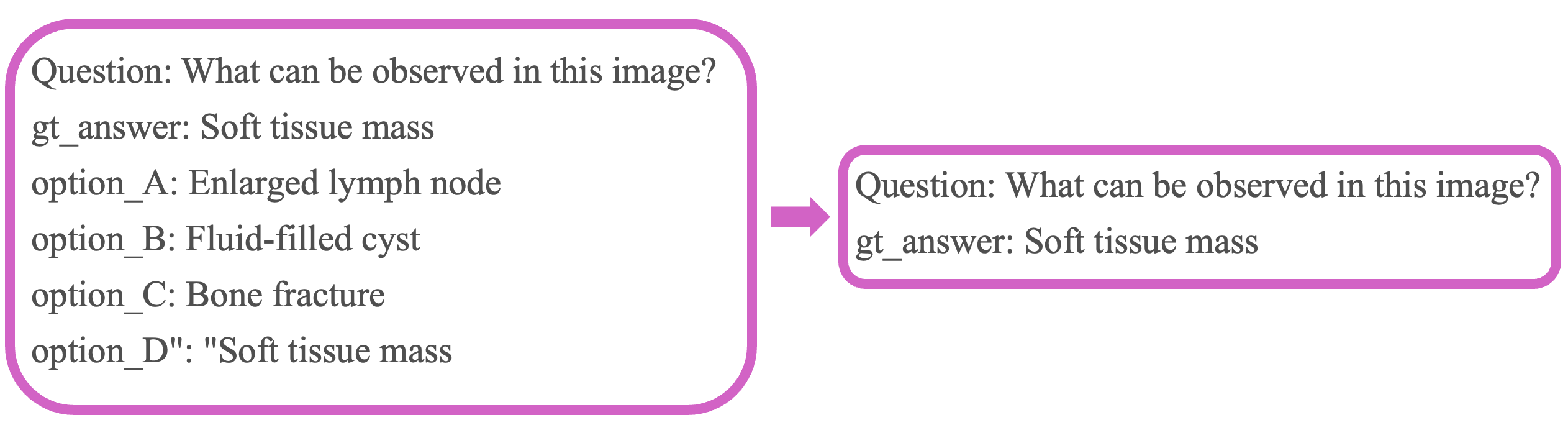}
    \caption{An example of question reformulation. The left side shows the original question-and-answer format in OmniMedVQA, while the right side displays the revised format used in our experiments. The gt\_answer represents the ground truth answer.}
    \label{revised_q}
\end{figure}

We use the OmniMedVQA dataset, which consists of medical images from multiple publicly available sources~\cite{hu2024omnimedvqa}. It includes CT, MRI, and X-ray images for VQA tasks. The original dataset features closed-ended question-answer pairs with multiple-choice answers.

For transparency, we only use images from publicly accessible datasets. The dataset contains 82,405 images and 88,996 QA pairs, which we split into a 70:30 training/testing ratio. We modify the questions to generate open-ended answers, removing the multiple-choice options and replacing the ground truth (answer numbers) with content, as shown in Figure~\ref{revised_q}. This adaptation allows us to evaluate the model’s ability to generate contextually relevant, open-ended responses for medical inquiries, ensuring its applicability to real-world clinical tasks where such responses are needed.

\subsection{Evaluation Metrics}
In our experiments, we use accuracy as the primary evaluation metric, aligning with the characteristics of the OmniMedVQA dataset. The dataset primarily consists of closed-ended questions (e.g., multiple-choice questions) based on medical images, enabling us to measure how often the model predicts the exact answer correctly. 

Although we adapted the dataset into open-ended questions, accuracy remains applicable due to the short, direct nature of the answers. This allows for a clear distinction between correct and incorrect responses, making accuracy a reliable performance indicator.

\subsection{Experimental Conditions}
We trained on our university's supercomputing facility with two NVIDIA 40 GB A100 GPUs to handle large-scale training. The process includes two components: BiomedCLIP and LLama3-8B. BiomedCLIP was trained with the same parameters as the original, ensuring consistent image encoding. For LLama3-8B, we used LoRA (Low-Rank Adaptation) \cite{hu2021lora} for efficient fine-tuning.

Key hyperparameters for LLama3-8B include:
\begin{itemize}
    \item Batch Size: 128
    \item Learning Rate: 0.0001, optimized with AdamW.
    \item LoRA Parameters: Alpha = 32, rank = 8 for efficient adaptation.
    \item Device and Data Types: bf16 for optimized memory, trained on CUDA.
\end{itemize}

We used gradient accumulation steps of 1, a cosine learning rate scheduler with 100 warmup steps, and saved checkpoints regularly with a custom FullModelMetaCheckpointer. The training prioritized accuracy and efficiency, fine-tuning both components to align image and text features.

\section{Results and Analysis}
In our evaluation on the OmniMedVQA dataset, the model was tested on 7,930 images and 8,832 question-answer (QA) pairs, achieving 73.4\% accuracy, as shown in Table~\ref{test_on_OmniMedVQA}. The model answered 6,487 questions correctly, with 3,441 open-ended and 3,046 yes/no questions correct. The test loss was 7.617, indicating potential for improved generalization.

\begin{table}[h] 
\centering 
\caption{Model performance on OmniMedVQA dataset} 
\label{test_on_OmniMedVQA}
\begin{tabular}{|c|c|c|} 
\hline 
\textbf{Question Type} & \textbf{Correct Answers} & \textbf{Incorrect Answers} \\ 
\hline 
Open-end Questions & 3,441 & 1,429 \\ 
\hline 
Yes/No Questions & 3,046 & 916 \\ 
\hline 
\textbf{Total} & 6,487 & 2,345 \\ 
\hline 
\end{tabular} 
\end{table}

\begin{figure}[htbp]
\centerline{\includegraphics[width=\linewidth]{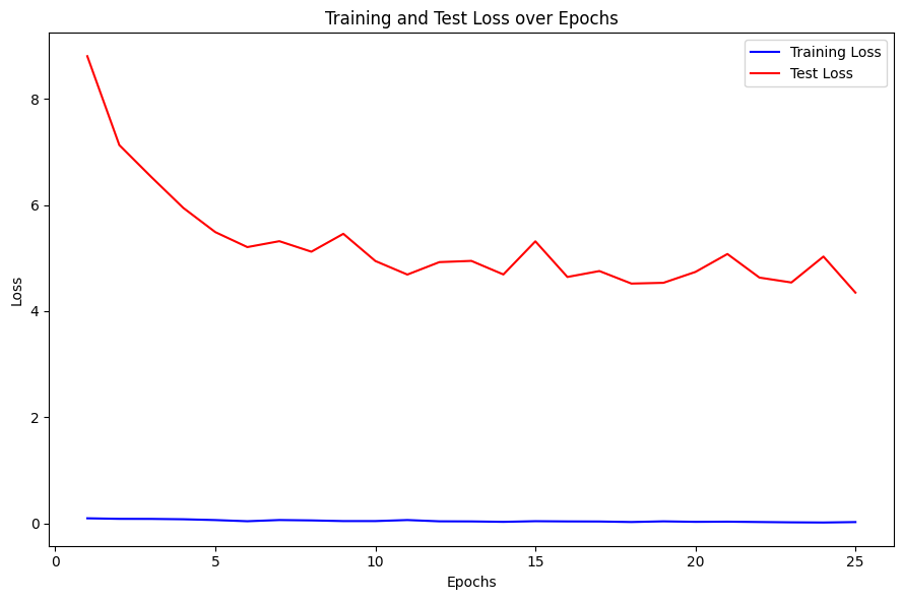}}
\caption{Training and test loss over epochs on OmniMedVQA.}
\label{best_loss_omnimedvqa}
\end{figure}

Figure~\ref{best_loss_omnimedvqa} shows the training and test loss trends. Both decrease rapidly in early epochs, but test loss remains higher than training loss.

Table~\ref{performance_multimodal} summarizes model performance across different imaging modalities. Microscopy images achieved the highest accuracy (78.5\%), followed by Ultrasound (77.2\%) and OCT (77.3\%). CT and X-ray images also performed well (75.8\% and 75.7\%, respectively), while MRI images had the lowest accuracy (69.2\%). The lower performance on MRI images may stem from their complexity, variability in scan types, and larger dataset size, which could challenge the model’s generalization.

\begin{table}[h!]
\centering
\caption{Performance across different modalities. X-Ray: X-Radiation; MRI: Magnetic Resonance Imaging;
OCT: Optical Coherence Tomography;
CT: Computed Tomography}
\label{performance_multimodal}
\begin{tabular}{|c|c|c|c|}
\hline
\textbf{Modalities} & \textbf{Total} & \textbf{Correct} & \textbf{Acc(\%)} \\ \hline
X-Ray               & 1562           & 1172             & 75.7             \\ \hline
Dermoscopy          & 1395           & 1000             & 72.4             \\ \hline
MRI                 & 6314           & 4325             & 69.2             \\ \hline
OCT                 & 925            & 709              & 77.3             \\ \hline
CT                  & 3144           & 2383             & 75.8             \\ \hline
Microscopy Images    & 1136           & 884              & 78.5             \\ \hline
Ultrasound          & 2185           & 1672             & 77.2             \\ \hline
Fundus Photography   & 1131           & 798              & 71.3             \\ \hline
\end{tabular}
\end{table}

\section{Discussion}
\subsection{Advantages of the Proposed Model}
\begin{table*}
\centering
\caption{Comparison of VQA Models. The last two columns show accuracy for different question types. N/A denotes unavailable data. B refers to billion.}
\label{comparison}
\resizebox{\textwidth}{!}{%
\begin{tabular}{|p{1.5cm}|p{1cm}|p{3cm}|p{2.5cm}|p{2cm}|p{3cm}|p{1.3cm}|p{1.3cm}|p{1.3cm}|}
\hline
\textbf{Reference} & \textbf{Years} & \textbf{Models} & \textbf{Dataset} & \textbf{Parameters} & \textbf{Resources} & \textbf{Open (\%)} & \textbf{Closed (\%)} & \textbf{Overall (\%)} \\ \hline
\cite{zhang2023biomedclip} & 2023 & ViT-B/16 + PubMedBERT & VQA-RAD & 13B & 16 $\times$ NVIDIA A100 GPUs & 67.0 & 76.5 & 72.7 \\ \hline
\cite{li2024llava} & 2023 & ViT-L/14 + LLaMa-7B & VQA-RAD & 13B & 8 $\times$ A100 GPUs & 61.5 & 84.2 & 75.2 \\ \hline
\cite{li2023masked} & 2023 & ViT-B/12 + BERT & VQA-RAD & N/A & 1 $\times$ Intel Xeon & 71.5 & 84.2 & 79.2 \\ \hline
\cite{hu2024omnimedvqa} & 2024 & BLIP-2 & OmniMedVQA & N/A & N/A & N/A & 48.12 & 48.12 \\ \hline
\cite{hu2024omnimedvqa} & 2024 & InstructBLIP & OmniMedVQA & N/A & N/A & N/A & 40.4 & 40.4 \\ \hline
\cite{hu2024omnimedvqa} & 2024 & RadFM & OmniMedVQA & N/A & N/A & N/A & 26.99 & 26.99 \\ \hline
\cite{chen2024huatuogpt} & 2024 & LLaVA-v1.5-LLaMA3-8B & OmniMedVQA & 34B & N/A & N/A & 76.7 & 76.7 \\ \hline
\hline
\textbf{Ours} & \textbf{2024} & \textbf{BiomedCLIP – LLaMA3-8B} & \textbf{Revised OmniMedVQA} & \textbf{8B} & \textbf{2 $\times$ A100 GPUs} & \textbf{70.7} & \textbf{76.9} & \textbf{73.4} \\ \hline
\end{tabular}%
}
\end{table*}

\begin{figure*}[htbp]
\centerline{\includegraphics[width=\textwidth]{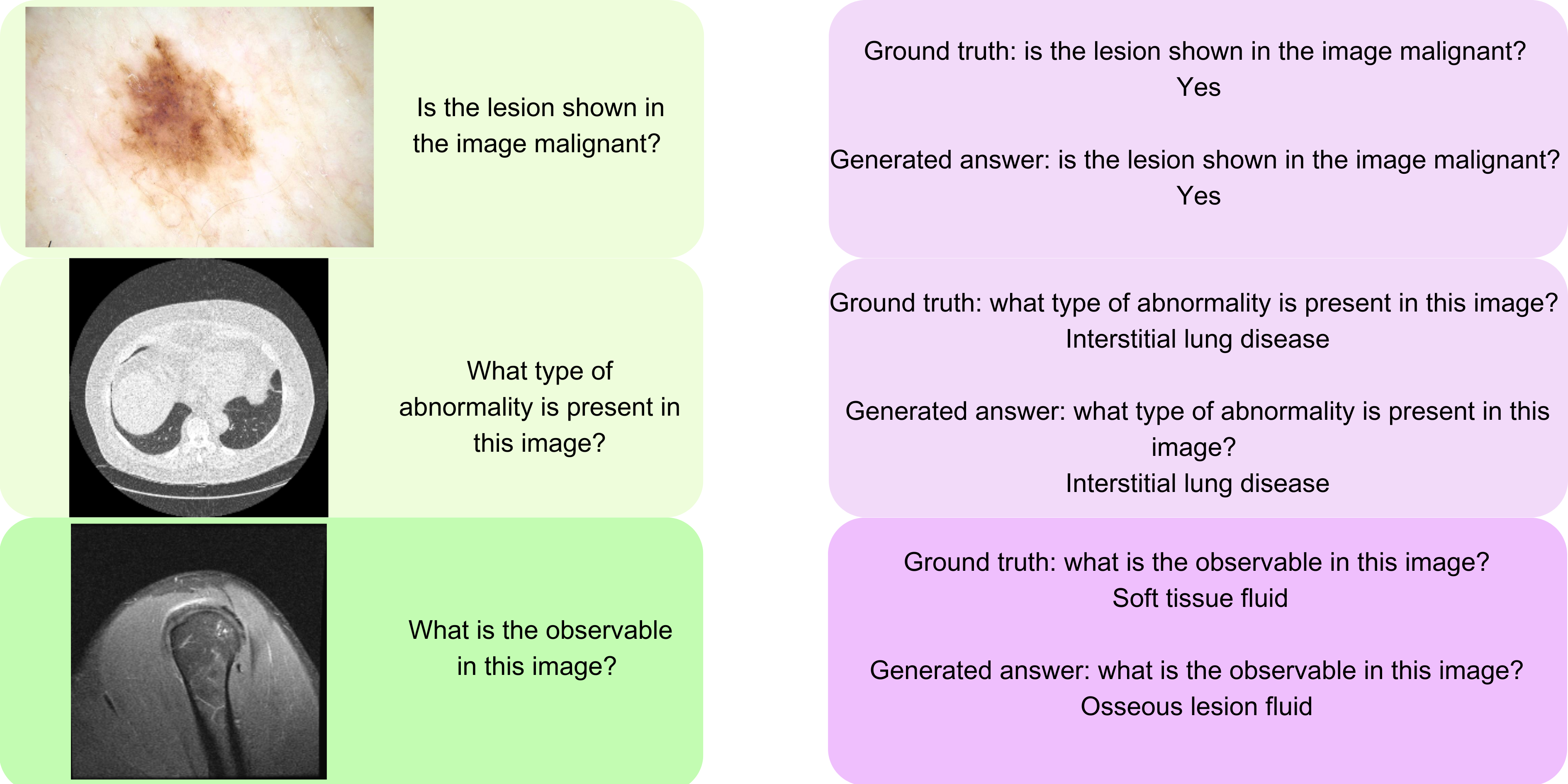}}
\caption{Model outputs for three distinct medical images. Light-colored modules indicate correct answers; dark-colored ones show errors.}
\label{case}
\end{figure*}

Table~\ref{comparison} compares our model with existing VQA models. A key advantage is our model's ability to handle open-ended questions, unlike models designed for closed-ended ones. While our model’s overall accuracy (73.4\%) is slightly lower than HuatuoGPT-Vision-34B (76.7\%), it outperforms it on closed-ended questions. Another strength is its efficiency: with only 8 billion parameters, our model runs on just two NVIDIA A100 GPUs, compared to HuatuoGPT-Vision-34B’s 34 billion parameters. This makes our model resource-efficient, cost-effective, and well-suited for real-world medical applications.

\subsection{Case Analysis}

Figure~\ref{case} illustrates our model's performance on three medical images. Correct answers are marked by light-colored modules, while incorrect answers are highlighted in dark-colored modules. The model correctly identifies a malignant lesion in a dermoscopic image, interstitial lung disease in a CT scan, and incorrectly identifies a shoulder MRI feature. Despite these successes, a limitation arises from dataset uniformity, where repetitive question-answer pairs may lead to overfitting, causing an `accuracy paradox' where performance appears better due to memorization rather than generalization. More varied training data is needed for robust model performance.

\section{Conclusion}
In this paper, we introduced a lightweight multimodal VQA model for medical imaging, combining BiomedCLIP for image feature extraction and LLaMA-3 for text encoding. Our model achieves state-of-the-art performance on the OmniMedVQA dataset, outperforming existing models with fewer computational resources, making it more suitable for resource-constrained clinical settings. Its ability to handle open-ended questions enhances its versatility for various medical tasks.

However, there are areas for future work. One limitation is the repetitive nature of some dataset questions, potentially causing an "accuracy paradox". Future efforts will focus on diversifying the dataset, improving generalization across modalities, extending the model to handle multi-step reasoning tasks, and enabling real-time inference for clinical support.

\bibliographystyle{splncs04}
\bibliography{MICCAI2025papertemplate}
\end{document}